\documentclass{CVM}

\graphicspath{{fig/}{figures/}}

\CVMsetup{
type      = {Research Article},
doi       = {},
title     = {Robust Global Structure-from-Motion via View Graph Pruning},
author    = {Jiamin Xu$^{1}$, Lixing Yao$^{1}$, Weichen Dai$^{1}$, Renshu Gu$^{1}$, Zunjie Zhu$^{1}$, Weiwei Xu$^{2}$, Gang Xu$^{1}$},
runauthor = {Jiamin Xu et al.},
runtitle  = {Robust Global SfM via Subgraph-guided View Graph Pruning},
abstract  = {
Structure-from-Motion (SfM) aims to estimate camera poses and reconstruct 3D structures from a collection of unordered images. Compared with incremental SfM, global SfM achieves better scalability by jointly estimating camera poses based on a view graph constructed from pairwise correspondences. However, its performance is highly sensitive to erroneous edges caused by visually ambiguous matches, which may lead to incorrect camera registration and reconstruction artifacts. In this work, we propose a subgraph-guided view graph pruning framework for robust global SfM. Our key idea is to exploit the internal consistency of reliable subgraphs to identify and remove unreliable connections. Specifically, we first partition the view graph into locally consistent subgraphs and perform global SfM within each subgraph to obtain reliable camera poses. We then apply RANSAC-based edge pruning across subgraphs to remove inconsistent edges, and finally perform global SfM on the refined view graph. Extensive experiments on ambiguous, sequential, and unordered image datasets demonstrate that our method improves the robustness of global SfM under challenging conditions. Further evaluation with neural rendering shows that the improved camera estimation leads to higher-quality novel view synthesis results.
},
keywords  = {Structure-from-Motion, Global SfM, View Graph, Graph Pruning, 3D Reconstruction},
copyright = {The Author(s)},
logo      = {logo},
}

\begin{document}

\maketitle

    \begin{figure}[b] \vskip -2mm
    \small\renewcommand\arraystretch{1.3}
        \begin{tabular}{p{80.5mm}} \toprule\\ \end{tabular}
        \vskip -4.5mm \noindent \setlength{\tabcolsep}{1pt}
        \begin{tabular}{p{3.5mm}p{80mm}}
    $1\quad $ & Hangzhou Dianzi University, Hangzhou, 310018, China.\\
    $2\quad $ & State Key Lab of CAD\&CG, Zhejiang University, Hangzhou, 310058, China.
    \end{tabular} \vspace {-3mm}
    \end{figure}

\section{Introduction}
\label{sec:intro}

Structure-from-Motion (SfM) is a fundamental computer vision technique that simultaneously estimates camera poses and reconstructs sparse 3D structures from multiple images. Over the past decade, SfM has found widespread applications across various domains, including indoor–outdoor 3D reconstruction~\cite{schoenberger2016mvs,yao2018mvsnet}, cultural heritage digitization~\cite{fuhrmann2014mve}, and, more recently, neural rendering~\cite{mildenhall2021nerf}.


\begin{figure}[t]
  \centering
  \includegraphics[width=1.0\linewidth]{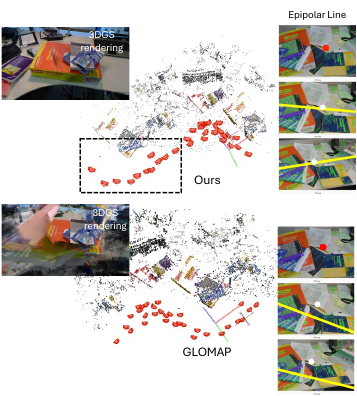}
  \caption{\textbf{Our view graph pruning strategy enhances the GLOMAP~\cite{pan2024glomap} method for scenes with visually ambiguous regions.} \emph{Right}: We evaluate epipolar geometry errors by marking a red point in the reference image and checking whether the epipolar line passes through the corresponding point (marked in white). \emph{Left}: We also assess novel view synthesis results using 3DGS~\cite{kerbl20233d}. The results show that our method produces higher-quality novel view images, demonstrating improved SfM consistency.}
  \label{fig:teaser}
\end{figure}

SfM methods can be broadly categorized into incremental and global approaches. Incremental methods start from an initial image pair and sequentially expand the reconstruction by adding new cameras and triangulated points, interleaving pose estimation, triangulation, and bundle adjustment~\cite{schonberger2016structure}. While they typically achieve high accuracy and robustness, their scalability is limited by repeated optimization. In contrast, global methods construct a view graph with pairwise correspondences and relative poses, and estimate all camera parameters simultaneously through rotation and translation averaging, offering much higher efficiency and scalability~\cite{pan2024glomap}.

However, because global SfM relies on the view graph as the foundation for averaging, its performance is highly sensitive to erroneous edges—incorrect pairwise relations caused by false feature matches. These edges are particularly common in scenes with repetitive structures, such as building facades or decorative patterns, where visually similar regions lead to ambiguous correspondences. Consequently, a key challenge in global SfM is to reliably identify and remove such spurious edges. One approach is to use RANdom SAmple Consensus (RANSAC)~\cite{fischler1981random} to filter out spurious edges. For instance, if the relative pose of an edge $a \leftrightarrow b$ does not align with other paths from $a$ through different nodes to $b$, it could be considered spurious. However, we found this method ineffective, as false consensus can occur. Specifically, the edge $a \leftrightarrow b$ may be correct but still mistakenly filtered out, while the path through other nodes to $b$ may contain a spurious edge, causing all paths to reach a false consensus.


As a result, we propose an enhanced strategy for identifying spurious edges. Our method first extracts a consensus subgraph and detects outlier edges across subgraphs that deviate from the dominant consensus. Specifically, we iteratively partition the cameras into subgraphs and perform global SfM within each subgraph. If intra-subgraph pose consistency—evaluated through loop consistency or by comparing two-view relative poses with the subgraph’s global estimates—is not satisfied, the subgraph is further subdivided. RANSAC is then applied across subgraphs to remove edges that violate inter-subgraph consistency. To resolve scale ambiguities between subgraphs, we employ the Perspective-n-Point (PnP) algorithm to estimate their relative poses. Finally, we re-run global SfM on the refined view graph to obtain the final reconstruction.

Our subgraph-based strategy differs fundamentally from the scene component reconstruction schemes used in GLOMAP~\cite{pan2024glomap}, COLMAP~\cite{schonberger2016structure}, and commercial software RealityScan~\cite{Realitycapture2016}. These methods reconstruct individual scene components independently and subsequently merge them into a complete model. In contrast, our subgraph partitioning is designed to filter out spurious edges from the view graph prior to global optimization. By eliminating unreliable connections early, we effectively prevent error propagation during rotation and translation averaging, leading to more accurate and geometrically consistent 3D reconstructions (Fig.~\ref{fig:teaser}).


Overall, our contributions can be summarized as:
\begin{itemize}

\item We propose a subgraph-guided view graph pruning framework that improves the robustness of global SfM under visually ambiguous conditions.

\item We introduce an iterative subgraph partitioning strategy based on intra-subgraph pose consistency, which identifies reliable local geometric structures from corrupted view graphs.

\item We develop an inter-subgraph edge pruning strategy based on RANSAC and relative pose consistency to remove unreliable connections while preserving valid geometric relationships.

\end{itemize}

We evaluate our SfM system on both sequential and unordered image datasets. Our method achieves competitive or superior camera-pose accuracy across diverse datasets, with particularly strong improvements over GLOMAP\cite{pan2024glomap} in visually ambiguous scenes. We also assess novel view synthesis results using 3DGS~\cite{kerbl20233d} (Fig.~\ref{fig:teaser}), demonstrating improved reconstruction accuracy. Regarding reconstruction efficiency, our method achieves comparable performance to existing global SfM techniques while being significantly faster than most incremental SfM approaches.

\section{Related Work}
\label{sec:related}

Structure-from-Motion mainly follows two paradigms: incremental and global.


\paragraph{Incremental and global SfM:} Incremental methods~\cite{agarwal2011building,frahm2010building,schonberger2016structure,snavely2006photo,wu2013towards} reconstruct scenes progressively, starting from an initial image pair and sequentially adding new images and 3D points while alternating between pose estimation, triangulation, and local or global bundle adjustment (BA). This strategy typically yields high accuracy and robustness. However, it often suffers from drift accumulation in large-scale reconstructions and is computationally expensive due to repeated BA operations. To alleviate these issues, several approaches perform distributed camera registration followed by global merging~\cite{cui2015efficient,zhu2018very,chen2020graph}. 

Global SfM methods~\cite{arie2012global,cai2021pose,cui2015global,pan2024glomap,wilson2014robust} estimate all camera poses simultaneously through rotation and translation averaging~\cite{cui2015global}, followed by triangulation and a single global bundle adjustment (BA). Recent advancements, such as GLOMAP~\cite{pan2024glomap}, enhance efficiency and scalability by jointly optimizing camera poses and 3D structures within a unified global optimization framework. Nevertheless, despite their efficiency, global SfM approaches remain sensitive to outliers, including incorrect correspondences and ambiguities arising from visually indistinguishable structures.

\label{sec:method}
\begin{figure*}[t!]
\centering
\includegraphics[width=1.0\linewidth]{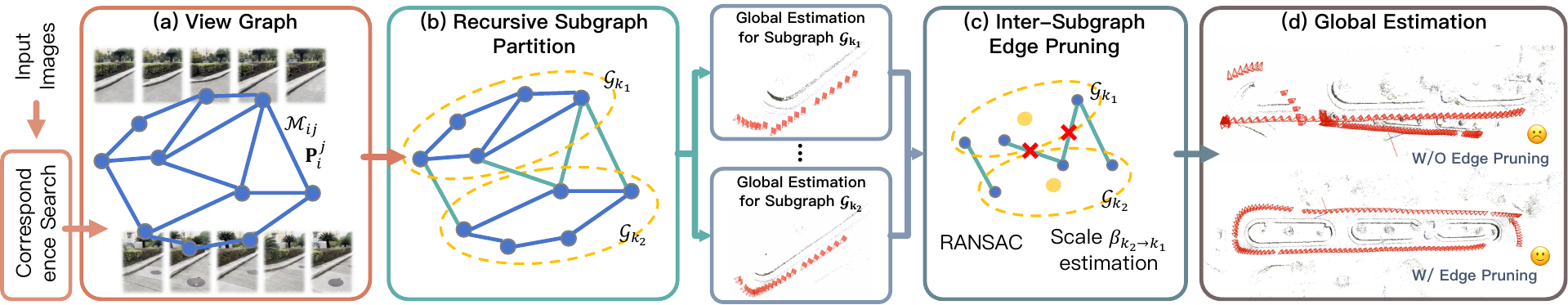}
\caption{\textbf{Our proposed pipeline.} Given a set of input images, our method: (a) performs correspondence search to construct a view graph; (b) recursively partitions the view graph into a set of internally consistent subgraphs and applies global SfM to each subgraph using GLOMAP\cite{pan2024glomap}; (c) performs inter-subgraph edge pruning to eliminate visually similar but geometrically inconsistent edges; and (d) applies global SfM to the pruned view graph to obtain the final reconstruction.}
\label{fig:pipeline} 
\end{figure*}


\paragraph{Outlier Rejection in SfM:}
A critical challenge in SfM is handling incorrect correspondences and erroneous camera pose estimates. Traditional approaches employ geometric verification using RANSAC~\cite{fischler1981random} with fundamental or essential matrix~\cite{andrew2001multiple} constraints to prune outliers. In incremental SfM, outliers can also be identified during triangulation~\cite{agarwal2011building} and bundle adjustment~\cite{triggs1999bundle} by measuring reprojection errors, removing points or images that deviate significantly from the reconstructed model~\cite{schonberger2016structure,kang2014robust}. Global SfM methods often incorporate robust optimization objectives (e.g., L1 norms or robust loss functions)~\cite{barron2019general} in rotation and translation averaging to reduce the impact of inconsistent pairwise estimates~\cite{wilson2014robust,goldstein2016shapefit}. These mechanisms are crucial for improving reconstruction stability; however, they rely heavily on the correctness of the underlying view graph. Consequently, a line of work focuses on improving the reliability and structure of the view graph.

\paragraph{View Graph Partition and Refinement:}
View-graph partitioning provides another line of work for improving the scalability and robustness of SfM. 
Bhowmick et al.~\cite{Bhowmick2014Divide} applied normalized cuts to the image match graph, reconstructed the resulting components independently, and aligned them using epipolar relationships between connecting images. 
Toldo et al.~\cite{Toldo2015Hierarchical} constructed a balanced hierarchical dendrogram over unordered images, enabling partial models to be created and merged in a bottom-up manner to reduce bundle-adjustment complexity and limit drift. 
Chen et al.~\cite{chen2020graph} proposed a graph-based parallel SfM framework that clusters images into strongly connected overlapping groups and merges local reconstructions through graph-based tree structures. 
Locher et al.~\cite{Locher2018Progress} introduced progressive SfM, which updates reconstructions as images arrive and avoids binding decisions made from incomplete view graphs. 
While these methods primarily use partitioning to decompose SfM into local reconstruction and merging stages, our approach employs partitioning to identify subgraphs and refine ambiguous inter-cluster edges, allowing the final model to be recovered directly by global SfM.

One approach to view-graph refinement is to directly remove incorrect edges. In a view-graph, nodes represent images and edges correspond to pairwise epipolar geometries, forming the structural backbone of global SfM. Early refinement methods focused on eliminating unreliable edges: Zach et al.~\cite{zach2010disambiguating,zach2008can} and Cui and Tan~\cite{cui2015global} exploited missing correspondences and loop-consistency checks to detect false geometric relations, while Roberts et al.~\cite{roberts2011structure} introduced an expectation–maximization framework incorporating temporal cues. Wilson and Snavely~\cite{wilson2013network} proposed a visibility-based local clustering measure to assess edge reliability, and learning-based approaches~\cite{cai2023doppelgangers, Xiang2025dopp++} formulated disambiguation as a binary classification task on image pairs. More recently, Wang et al.~\cite{wang2024tc} leveraged high-level spatial context through a novel track-community structure to segment the scene, detect and correct ambiguous regions, and perform partial reconstructions and merging, thereby improving reconstruction accuracy in visually ambiguous scenes.

Our method also focuses on view-graph refinement in visually ambiguous scenes. Unlike previous approaches, we enhance global SfM by applying RANSAC-based edge refinement between subgraphs (clusters), producing reconstruction results directly from global SfM rather than by merging partial reconstructions. Leveraging global SfM in this way makes our method achieve a favorable trade-off between efficiency and robustness.

\section{Proposed Method}

\subsection{Overview} 

Our approach builds upon the GLOMAP framework~\cite{pan2024glomap}. As shown in Fig.~\ref{fig:pipeline}, we first reconstruct the view graph using the original GLOMAP pipeline (Sec.~\ref{sec:preliminaries}). The view graph is then iteratively partitioned into subgraphs until intra-camera pose consistency is achieved across all subgraphs (Sec.~\ref{sec:subgraph_partition}). For each resulting subgraph, we perform global estimation, including rotation averaging, global positioning, and bundle adjustment. Next, inter-subgraph edge pruning is applied to remove inconsistent connections from the graph (Sec.~\ref{sec:view_graph_filtering}). Finally, global estimation is re-applied on the refined view graph to produce the final reconstruction. 

In the following sections, we briefly review the key components of GLOMAP as preliminaries (Sec.~\ref{sec:preliminaries}), and then introduce the subgraph partitioning and inter-subgraph edge pruning components of our method.


\subsection{Preliminaries}
\label{sec:preliminaries}

\paragraph{View graph reconstruction.}

In the first step, GLOMAP~\cite{pan2024glomap} performs feature extraction and matching, followed by verification through two-view geometry estimation. The remaining correspondences are considered inliers. Based on these inliers, a view graph $\mathcal{G} = \{\mathcal{V}, \mathcal{E}\}$ is constructed, where $\mathcal{V}$ denotes the set of images and $\mathcal{E}$ denotes the set of edges. 

An edge $\langle i, j \rangle \in \mathcal{E}$ indicates that the corresponding image pair contains a sufficient number of matching inlier points $\mathcal{M}_{ij} = \{ \langle\mathbf{x}^i_n, \mathbf{x}^j_n\rangle | n = 1, \dots, M_{ij} \}$ where $\mathbf{x}^i_n, \mathbf{x}^j_n \in \mathbb{R}^2$ denote the 2D pixel coordinates of the $n$-th matching point in images $i$ and $j$, respectively. A reliable relative pose $\mathbf{P}^j_i = (\mathbf{R}^j_i, \mathbf{t}^j_i)$ can then be estimated from their two-view geometry~\cite{PoseLib}, where $\mathbf{P}^j_i$ denotes the pose of image~$j$ expressed in the coordinate of image~$i$.

Next, using the view graph as input, the global camera pose estimation step simultaneously estimates absolute camera poses $\hat{\mathbf{P}}_i \in \mathrm{SE(3)}$. The hat notation $\hat{\cdot}$ is used to distinguish the globally estimated poses from the relative poses in the view graph obtained via two-view geometry. This procedure involves rotation averaging and global positioning to estimate camera orientations and positions, respectively, followed by a global bundle adjustment to refine the entire reconstruction.

\paragraph{Rotation averaging.}

Rotation averaging aims to estimate the absolute rotations $\{\hat{\mathbf{R}}_1, ..., \hat{\mathbf{R}}_N\}$ from a set of relative rotations. GLOMAP~\cite{pan2024glomap} formulates this task as a nonlinear optimization problem:
\begin{equation}  
\operatorname*{argmin}_{\hat{\mathbf{R}}_1, ..., \hat{\mathbf{R}}_N} 
\sum_{\langle i, j \rangle \in \mathcal{E}} 
\mathrm{d}(\mathbf{R}^j_i,\hat{\mathbf{R}}_j^\top \hat{\mathbf{R}}_i),
\end{equation}
where $\mathrm{d}(\cdot)$ denotes a distance metric on $\mathrm{SO(3)}$. Since this optimization is inherently nonlinear and difficult to solve directly, GLOMAP adopts a linearization strategy~\cite{chatterjee2013efficient}.

\paragraph{Global positioning.}

Next, instead of performing translation averaging followed by global triangulation, GLOMAP~\cite{pan2024glomap} estimates camera positions and 3D points jointly through a unified optimization. The formulation discards relative translation constraints and relies solely on camera-ray consistency:
\begin{equation} 
\operatorname*{argmin}_{\mathbf{X},\,\mathbf{c},\,s_{ik}}
\sum_{i,k}\rho\left(\left\| \mathbf{v}_{ik} - s_{ik}(\mathbf{X}_k - \mathbf{c}_i) \right\|_2^2\right),
\quad \text{s.t. } d_{ik} \ge 0, 
\end{equation}
where $\mathbf{v}_{ik}$ is the camera ray that observes the 3D point $\mathbf{X}_k$ from the camera center $\mathbf{c}_i$, and $s_{ik}$ is a scale factor. The function $\rho(\cdot)$ denotes the Huber loss~\cite{huber1992robust}, which serves as a robustifier.

\subsection{Subgraph Partition}
\label{sec:subgraph_partition}

\begin{figure}[t!]
  \centering
  \includegraphics[width=1.0\linewidth]{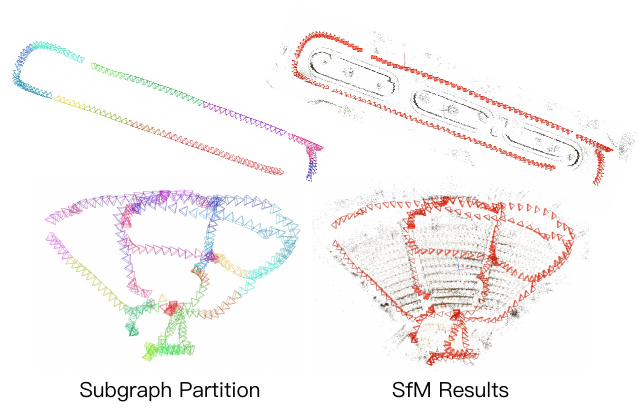}
  \caption{\textbf{Visualization of subgraph partition.} Each subgraph in the view graph is shown in a different color, demonstrating that our partitioning algorithm can separate visually similar views into distinct subgraphs.}
  \label{fig:subgraphs}
\end{figure}

To identify spurious edges in $\mathcal{G}$, we first partition it into a set of subgraphs ${\mathcal{G}_1, \dots, \mathcal{G}_K}$, as shown in Fig.~\ref{fig:subgraphs}. This partitioning is performed using the Louvain community detection algorithm, which greedily optimizes modularity to identify subgraphs with dense intra-community connections and sparse inter-community connections~\cite{blondel2008louvain} and is applied recursively until intra-subgraph camera pose consistency is achieved. After each partition, we perform GLOMAP's global estimation step for each subgraph to obtain the camera poses $\hat{\mathbf{P}}_{i} = (\hat{\mathbf{R}}_{i}, \hat{\mathbf{t}}_{i}) \in \mathrm{SE(3)}$, which are then used to determine whether further subdivision is necessary and will also be used in subsequent edge pruning step.

Specifically, for a subgraph $\mathcal{G}_k$ with $n$ images (initially $\mathcal{G}$), we compute the similarity matrix $\mathbf{W} \in \mathbb{R}^{n \times n}$ for all image pairs. Each similarity is defined by the number of inlier correspondences $|\mathcal{M}_{ij}|$. We regard $\mathbf{W}$ as the weighted adjacency matrix of $\mathcal{G}_k$, where $\mathbf{A}_{ij}=\mathbf{W}_{ij}$. The subgraph is then partitioned using the Louvain community detection algorithm~\cite{blondel2008louvain}, which greedily optimizes modularity to discover densely connected communities.

For a weighted view graph, the modularity of a partition is defined as
\begin{align}
\mathbf{Q} = \frac{1}{2m} \sum_{i,j}
\left[
\mathbf{A}_{ij} - \frac{k_i k_j}{2m}
\right]
\delta(c_i,c_j),
\end{align}
where $\mathbf{A}_{ij}$ denotes the edge weight between images $i$ and $j$, 
$k_i=\sum_j \mathbf{A}_{ij}$ is the weighted degree of node $i$, 
$m=\frac{1}{2}\sum_{i,j}\mathbf{A}_{ij}$ is the total edge weight of the graph, 
$c_i$ is the community assignment of node $i$, and $\delta(c_i,c_j)$ equals 1 if $i$ and $j$ belong to the same community and 0 otherwise.

The Louvain algorithm starts by assigning each image to an individual community. 
For each node $i$, it evaluates the modularity gain obtained by moving $i$ into the community of one of its neighboring nodes. 
The modularity gain is computed as
\begin{align}
\Delta \mathbf{Q} =
\left[
\frac{\Sigma_{\mathrm{in}} + k_{i,\mathrm{in}}}{2m}
-
\left(
\frac{\Sigma_{\mathrm{tot}} + k_i}{2m}
\right)^2
\right]
\notag
\\
-
\left[
\frac{\Sigma_{\mathrm{in}}}{2m}
-
\left(
\frac{\Sigma_{\mathrm{tot}}}{2m}
\right)^2
-
\left(
\frac{k_i}{2m}
\right)^2
\right],
\end{align}
where $\Sigma_{\mathrm{in}}$ is the sum of edge weights inside the target community, 
$\Sigma_{\mathrm{tot}}$ is the sum of edge weights incident to nodes in the target community, 
$k_i$ is the weighted degree of node $i$, and $k_{i,\mathrm{in}}$ is the sum of edge weights from node $i$ to nodes in the target community.

Node $i$ is moved to the neighboring community that yields the largest positive $\Delta \mathbf{Q}$; otherwise, it remains in its current community. 
This local optimization is repeated until no individual node movement can further increase modularity. 
After that, each discovered community is collapsed into a super-node, and the edge weights between super-nodes are obtained by summing the weights of edges between the corresponding communities. 
The local optimization and community aggregation steps are then repeated on the coarsened graph until the modularity no longer increases. 
The final communities are used as the partitioned subgraphs for subsequent inter-cluster edge refinement.

\paragraph{Intra-subgraph camera pose consistency.}

As a termination criterion for the subdivision, we assess the consistency of camera poses within each subgraph using two metrics. The first is loop consistency, and the second is relative-to-absolute camera pose consistency, which measures the agreement between relative poses estimated from two-view geometry and absolute camera poses obtained from the subgraph’s global estimation.

For \textbf{loop consistency}, a subgraph is considered inconsistent if any loop—for example, $\mathbf{R}^i_p\mathbf{R}^p_j\mathbf{R}^j_i$—deviates significantly from the identity rotation. Since the relative translations do not share a consistent scale, we only evaluate the loop rotation error:
\begin{align}
e_\mathrm{l} = \| \mathrm{Log}(\mathbf{R}^i_j\dots\mathbf{R}^j_i) \|_2
\end{align}
where $\mathrm{Log}(\cdot)$ maps a rotation matrix in $\mathrm{SO(3)}$ directly to its corresponding axis-angle vector. For efficiency, we only consider loops of length $l$ that contain fewer than eight cameras. Since consistency is evaluated within each subgraph, this simplification is considered acceptable. If any loop exhibits a loop-consistency error $e_\mathrm{l}$ greater than 
$5^\circ \cdot \left\lfloor \frac{l - 1}{2} \right\rfloor$, 
the subgraph is further subdivided.

The \textbf{relative-to-absolute camera pose consistency} metric evaluates the agreement between the relative pose of an edge in the view graph and the corresponding pose derived from the subgraph’s global estimation in GLOMAP. Since these poses may have different scales, only the rotation error and the scale-invariant translation error are considered:
\begin{align}
e_\mathrm{r} &= \| \mathrm{Log}(\hat{\mathbf{R}}_{j}\mathbf{R}^j_i\hat{\mathbf{R}}_{i}^\top) \|_2,\\
e_\mathrm{t} &=
\mathrm{acos}\left(
\frac{
(\mathbf{t}^j_i)^\top
\hat{\mathbf{R}}_j^\top
(\hat{\mathbf{t}}_{i} - \hat{\mathbf{t}}_{j})
}{
\|\mathbf{t}^j_i\|\,
\left\|
\hat{\mathbf{R}}_j^\top
(\hat{\mathbf{t}}_{i} - \hat{\mathbf{t}}_{j})
\right\|
}
\right).
\end{align}
If, for any edge in the subgraph, either $e_\mathrm{r}$ or $e_\mathrm{t}$ exceeds $10^\circ$, the subgraph is further subdivided.


\subsection{Inter-Subgraph Edge Pruning}
\label{sec:view_graph_filtering}

After partitioning the entire view graph into a set of subgraphs, we perform inter-subgraph edge pruning to remove inconsistent edges from $\mathcal{G}$ (Fig.~\ref{fig:pipeline}). 
The input to this step consists of the complete view graph $\mathcal{G}$ and the set of subgraphs $\{\mathcal{G}_1, \dots, \mathcal{G}_K\}$. 
For the complete view graph, no global estimation is performed; consequently, it does not contain absolute camera poses or a global point cloud. In contrast, after the subgraph partitioning step, each subgraph $\mathcal{G}_k$ maintains its own set of absolute camera poses $\hat{\mathbf{P}}^k_i = (\hat{\mathbf{R}}^k_i, \hat{\mathbf{t}}^k_i) \in \mathrm{SE(3)}$ and a corresponding local point cloud $\mathbf{X}^k$. Here, the subscript $k$ indicates that both the camera poses and the point cloud are represented in the local coordinate frame.

For any two subgraphs $\mathcal{G}_{k_1}$ and $\mathcal{G}_{k_2}$, 
we define the set of inter-subgraph edges as those in the view graph whose endpoints belong to different subgraphs: $\{ \langle i,j \rangle \in \mathcal{E}' | i\in \mathcal{G}_{k_1}, j\in \mathcal{G}_{k_2} \}
$. Since the absolute poses of different subgraphs are estimated in their own local coordinate systems and may differ in scale, for each image $j$ on an inter-subgraph edge, we use the Perspective-n-Point (PnP) method to estimate its pose in the coordinate system of image $i$’s subgraph $\mathcal{G}_{k_1}$. 
We then align the two subgraphs by estimating a relative scale factor $\beta_{k_2 \to k_1}$ based on all edges connecting the two subgraphs.

For an image $i$ in subgraph $\mathcal{G}_{k_1}$, we obtain its 2D correspondences with image $j$ as $\langle \mathbf{x}_i, \mathbf{x}_j \rangle$. To simplify notation, we slightly overload the symbol $\mathbf{x}_i$ to represent all matching points in image $i$ for the image pair $\langle i,j \rangle$. The corresponding 3D points can be retrieved from the point clouds of $\mathcal{G}_{k_1}$ and $\mathcal{G}_{k_2}$, denoted as $\mathbf{X}_i^{k_1}$ and $\mathbf{X}_j^{k_2}$, respectively. The camera poses $\hat{\mathbf{P}}^{k_1}_i$ and $\hat{\mathbf{P}}^{k_2}_j$ are also assumed to be known.

The camera pose of image $j$ in the local coordinate system of $\mathcal{G}_{k_1}$, denoted as $\hat{\mathbf{P}}^{k_1}_j$, can be estimated using the PnP formulation:
\begin{equation}
s \begin{bmatrix} \mathbf{x}_j \ 1 \end{bmatrix}^\top = \mathbf{K}_j(\hat{\mathbf{P}}^{k_1}_j)^{-1} \begin{bmatrix} \mathbf{X}^{k_1}_j \ 1 \end{bmatrix}^\top,
\end{equation}
where $\mathbf{K}_j$ denotes the camera intrinsic matrix obtained in the previous step. To solve this problem, we employ the Efficient PnP (EPnP) algorithm~\cite{lepetit2009ep}, which introduces a set of virtual control points to linearize the otherwise nonlinear PnP problem, thereby significantly improving computational efficiency. 

For inter-subgraph edges, we apply RANSAC~\cite{fischler1981random} along the path connecting $\mathcal{G}_{k_1}$ and $\mathcal{G}_{k_2}$. In each RANSAC iteration, a scale factor $\beta_{k_2 \to k_1}$ is estimated to align the scales between the two subgraphs, as their reconstructions are generally inconsistent in scale. First, we define the pose
of each subgraph, as the average pose of all images in subgraph as $\mathbf{P}_{c_1}^{k_1}$ and $\mathbf{P}_{c_2}^{k_2}$, in each local coordinate. Then we can compute the relative pose between ${c_1}$ and ${c_2}$ in the $\mathcal{G}_{k_2}$'s local coordinate:
\begin{equation}
\dot{\mathbf{P}}^{k_2}_{c_1 \to i \to j \to c_2} = {(\mathrm{S}_{k_1}(\hat{\mathbf{P}}_{c_2}^{k_2}}))^{-1} \hat{\mathbf{P}}_j^{k_2}({\hat{\mathbf{P}}_j^{k_1}})^{-1}{\hat{\mathbf{P}}_{c_1}^{k_1}}.
\end{equation}
Here, $\dot{\mathbf{P}}^{k_2}$ denotes the pose in the coordinate system of $\mathcal{G}{k_2}$ after scaling to align with $\mathcal{G}{k_1}$, and $\mathrm{S}_{k_1}(\cdot)$ represents the operation of scaling a subgraph to match $\mathcal{G}_{k_1}$:
\begin{align}
\mathrm{S}_{k_1}(\hat{\mathbf{P}}_{c_2}^{k_2})= \begin{bmatrix} \hat{\mathbf{R}}^{k_2}_{c_2} & \beta_{k_2 \to k_1} \hat{\mathbf{t}}^{k_2}_{c_2} \\ 0 & 1 \end{bmatrix}, 
\end{align}

To determine the scale factor, we minimize the discrepancies among the translation differences of all relative poses $\mathbf{P}^{k_2}_{c_1 \to j \to c_2}$ for edges connecting subgraphs $\mathcal{G}_{k_1}$ and $\mathcal{G}_{k_2}$:
\begin{align}
\beta^* &= \operatorname*{argmin}_{\beta} \sum_{\langle i,j \rangle \in \mathcal{E}''} \| \dot{\mathbf{t}}^{k_2}_{c_1 \to i \to j \to c_2} - \bar{\mathbf{t}}^{k_2}_{c_1 \to c_2} \|_2^2 ,
\\
\bar{\mathbf{t}}^{k_2}_{c_1 \to c_2} &= \frac{1}{|\mathcal{E}''|}\sum_{\langle i,j \rangle \in \mathcal{E}''}\dot{\mathbf{t}}^{k_2}_{c_1 \to i \to j \to c_2},
\end{align}
where $\dot{\mathbf{t}}^{k_2}$ denotes the translation component of $\dot{\mathbf{P}}^{k_2}$, and $\mathcal{E}''$ represents the subset of edges sampled from the original set $\mathcal{E}'$ connecting the two subgraphs during the RANSAC procedure. Since this is a linear least-squares problem, it can be solved directly.

After scaling, we calculate the rotation error for each sampled edge in $\mathcal{E}''$:
\begin{align}
E_\mathrm{r} &= \| \mathrm{Log}(\dot{\mathbf{R}}^{k_2}_{c_1 \to i \to j \to c_2}) - \bar{\mathbf{r}}^{k_2}_{c_1 \to c_2} \|_2, \\
\bar{\mathbf{r}}^{k_2}_{c_1 \to c_2} &= \frac{1}{|\mathcal{E}''|}\sum_{\langle i,j \rangle \in \mathcal{E}''}\mathrm{Log}(\dot{\mathbf{R}}^{k_2}_{c_1 \to i \to j \to c_2}),
\end{align}
and the translation error:
\begin{align}
E_\mathrm{t} &= \| \dot{\mathbf{t}}^{k_2}_{c_1 \to i \to j \to c_2} - \bar{\mathbf{t}}^{k_2}_{c_1 \to c_2} \|_2.
\end{align}

During each RANSAC iteration, we sample $0.4\cdot|\mathcal{E}'|$ edges between each pair of subgraphs. The thresholds for $E_\mathrm{r}$ and $E_\mathrm{t}$ are set to $5^\circ$ and $0.05\cdot \left\| \bar{\mathbf{t}}^{k_2}_{c_1 \to c_2} \right\|_2$, respectively. The RANSAC procedure runs for 30 iterations, during which we retain the largest subset $\mathcal{E}''$. The remaining edges, i.e., $\mathcal{E}' \setminus \mathcal{E}''$, are then discarded. After pruning the inter-subgraph edges, global estimation is reapplied on the updated view graph to obtain the final reconstruction, as shown in Fig.~\ref{fig:pipeline}.

\begin{figure*}[t!]
\centering
\includegraphics[width=0.95\linewidth]{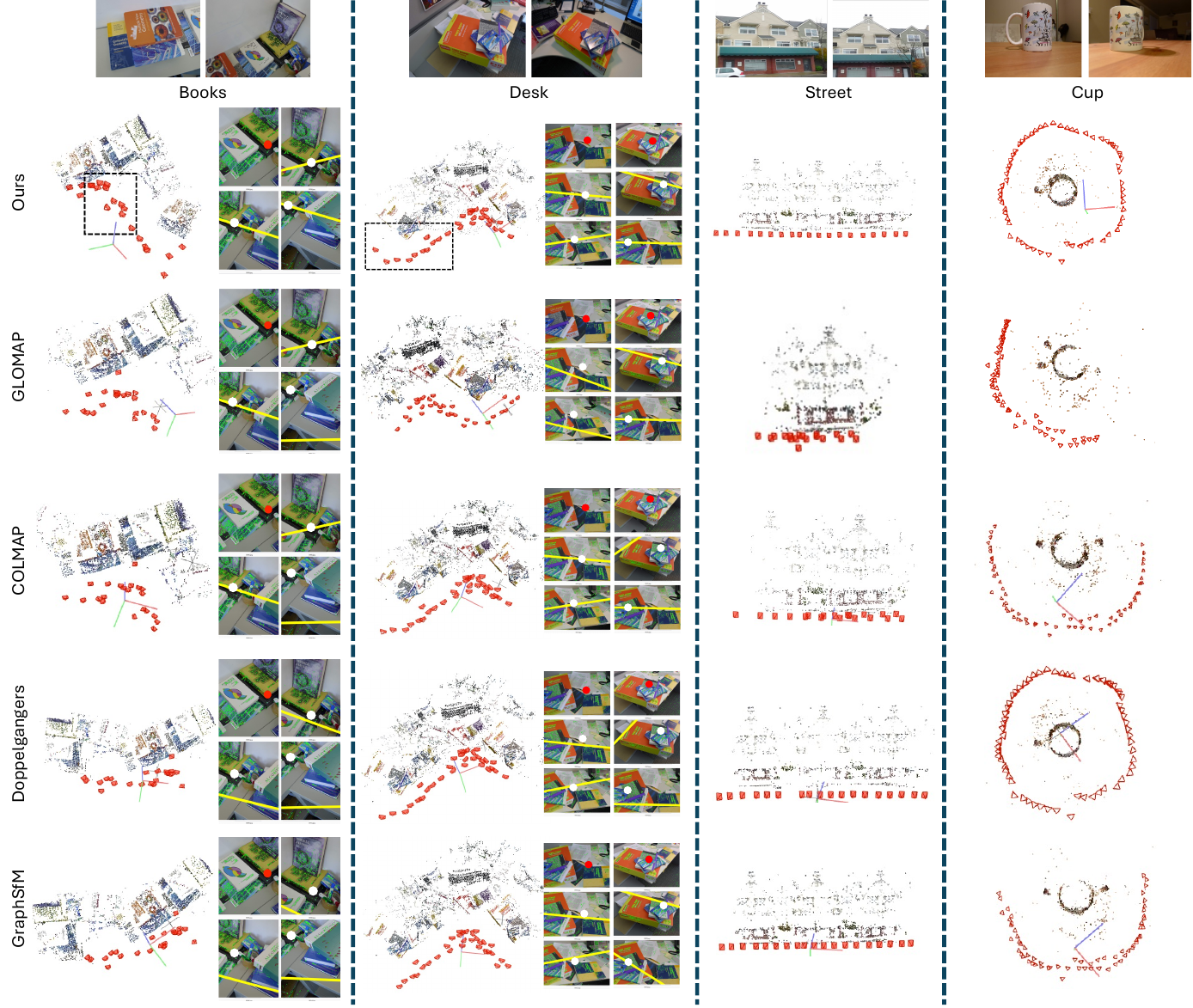}
\caption{\textbf{Qualitative evaluation on ambiguous image datasets.} \emph{The first row} shows the input images of selected views. \emph{The second row} shows our results and the rest rows shows competing methods. In each unit, we provide the SfM results (point cloud and camera trajectory) as well as epipolar line verification. The top left corner of the book is selected (marked in red) and epipolar lines are drawn on the other three views; ideally the epipolar line should pass through the corresponding point (marked in white) in every view.}
\label{fig:cmp1} 
\end{figure*}

\section{Experiments}

We compare our method with five SfM techniques, evaluating performance both quantitatively and qualitatively. Among these methods, GLOMAP~\cite{pan2024glomap} is a global SfM approach and serves as our baseline, as our method is based on it. COLMAP~\cite{schonberger2016structure} is the most widely used incremental SfM method and is often employed as the first step in neural rendering pipelines such as 3DGS~\cite{kerbl20233d}. GraphSfM~\cite{chen2020graph} is an SfM method designed to handle large-scale scenes. Recent learning-based SfM methods Doppelgangers~\cite{cai2023doppelgangers} and Doppelgangers++~\cite{Xiang2025dopp++} are designed to handle scenes with ambiguous structures.

\paragraph{Datasets.}

We evaluate our method on four types of datasets: ambiguous image datasets~\cite{roberts2011structure,jiang2012seeing}, sequential image datasets~\cite{knapitsch2017tanks}, unordered Internet image datasets~\cite{heinly2014correcting}, and our captured dataset. For all evaluated scenes, the ground-truth camera poses are reconstructed using RealityScan~\cite{Realitycapture2016} with manually annotated control points. To ensure the reliability of the reconstructed poses, we further conduct epipolar line verification on the resulting camera geometry.



\subsection{Evaluation on Ambiguous Image Datasets}

We test our method on the datasets from \cite{roberts2011structure}, specifically the ``books'', ``desk'', ``street'' and ``cup'' benchmark datasets. The visualization results are shown in Fig. \ref{fig:cmp1}. For every unit, we show the reconstructed results as well as epipolar line verification on the right. We select a reference point on the first image (marked in red), and draw the epipolar lines in the other 3 views, which should ideally pass through the corresponding point (marked in white). Our method shows robust performance on all four datasets.

GLOMAP \cite{pan2024glomap} occasionally gives reasonable results for ambiguous datasets such as ``Books''. However, in some cases, such as ``desk'', GLOMAP incorrectly registers cameras and points of duplicated structures, resulting in incorrect reconstruction. Additionally, GLOMAP performs poorly on the ``cup'' dataset, where it mistakenly connects the front and back faces of the cup in the view graph, due to their visual similarity, resulting in suboptimal performance.
COLMAP \cite{schonberger2016structure} produces incomplete structures for the ``cup'' dataset, although slightly better than GLOMAP. For ``books'' and ``desk'', the epipolar line verification shows that COLMAP is inaccurate as the epipolar lines do not always pass through the top left corner of the book (marked in white). 
GraphSfM \cite{chen2020graph} cannot reconstruct the complete camera trajectory for ``cup''. It also does not satisfy the epipolar constraints well. Doppelgangers~\cite{cai2023doppelgangers} performs well on the ``cup'' dataset but still struggles with registration accuracy in other cases. In contrast, all four scenes are successfully recovered using our proposed method.

Table \ref{tab:comparison} reports the quantitative comparison on four ambiguous image datasets, i.e., Books, Desk, Street, and Cup. For all evaluations, we report the AUC (Area Under the recall Curve) scores calculated from the maximum of relative rotation and translation error between every image pair, similar to~\cite{pan2024glomap}, such an error formulation considers the deviation between every possible camera pair.
We evaluate all methods using the AUC thresholds of $3^\circ$, $5^\circ$, and $10^\circ$. On Books, our method achieves the best results across all thresholds, reaching 98.7, 99.3, and 99.4, which significantly outperforms the second-best Doppelgangers++~\cite{Xiang2025dopp++} by 54.2, 51.3, and 48.7 points, respectively. On Desk, our method also consistently ranks first, improving over the second-best COLMAP by 9.9, 8.4, and 7.1 points. For Street, our method obtains 96.0, 97.6, and 98.8, again achieving the best performance and surpassing Doppelgangers++~\cite{Xiang2025dopp++} at all thresholds. These results demonstrate that our method is particularly effective in resolving ambiguous epipolar relationships and producing more reliable image associations.

On Cup, Doppelgangers++~\cite{Xiang2025dopp++} achieves the best performance, while our method ranks second with 66.4, 78.4, and 87.6. Although our method is not the best on this dataset, it still clearly outperforms COLMAP, GLOMAP, Doppelgangers~\cite{cai2023doppelgangers}, and GraphSfM under all three thresholds. Overall, across the four datasets, our method achieves the highest average AUC values of 90.1, 93.7, and 96.4 at $3^\circ$, $5^\circ$, and $10^\circ$, respectively. This indicates that our method provides strong and stable performance on ambiguous image datasets, especially in challenging cases where repetitive structures can easily lead to incorrect geometric verification.

Since metrics such as AUC cannot fully capture the accuracy of camera registration, we also evaluate our method for novel view synthesis using 3DGS~\cite{kerbl20233d} in scenes without ground truth camera poses. Table \ref{tab:3dgs} presents the novel view synthesis evaluation results. We use $80\%$ of the images for training and $20\%$ for testing, and all models are trained for 30,000 iterations. We compare the Peak Signal-to-Noise Ratio (PSNR) on the testing images, and the results show that our method achieves competitive performance and consistently improves over traditional SfM approaches, demonstrating its ability to produce globally consistent camera poses.

\begin{table*}[t]
\centering
\setlength\tabcolsep{2.0pt}
\resizebox{\linewidth}{!}{
\begin{tabular}{l *{18}{c}}
\hline
\multirow{2}{*}{Method} 
& \multicolumn{3}{c}{COLMAP\cite{schonberger2016structure}} 
& \multicolumn{3}{c}{GLOMAP\cite{pan2024glomap}} 
& \multicolumn{3}{c}{Doppelgangers\cite{cai2023doppelgangers}} 
& \multicolumn{3}{c}{Doppelgangers++\cite{Xiang2025dopp++}} 
& \multicolumn{3}{c}{GraphSfM\cite{chen2020graph}} 
& \multicolumn{3}{c}{Ours} \\ 
\cmidrule(lr){2-4}
\cmidrule(lr){5-7}
\cmidrule(lr){8-10}
\cmidrule(lr){11-13}
\cmidrule(lr){14-16}
\cmidrule(lr){17-19}
& {$auc@3^\circ\uparrow$} & {$auc@5^\circ\uparrow$} & {$auc@10^\circ\uparrow$} 
& {$auc@3^\circ\uparrow$} & {$auc@5^\circ\uparrow$} & {$auc@10^\circ\uparrow$} 
& {$auc@3^\circ\uparrow$} & {$auc@5^\circ\uparrow$} & {$auc@10^\circ\uparrow$} 
& {$auc@3^\circ\uparrow$} & {$auc@5^\circ\uparrow$} & {$auc@10^\circ\uparrow$} 
& {$auc@3^\circ\uparrow$} & {$auc@5^\circ\uparrow$} & {$auc@10^\circ\uparrow$} 
& {$auc@3^\circ\uparrow$} & {$auc@5^\circ\uparrow$} & {$auc@10^\circ\uparrow$}  \\ 
\hline
Books~\cite{roberts2011structure} 
& 18.2 & 27.4 & 38.8 
& 17.3 & 27.0 & 40.0 
& 43.5 & 47.4 & 50.4 
& \underline{44.5} & \underline{48.0} & \underline{50.7} 
& 20.2 & 28.8 & 39.5 
& \textbf{98.7} & \textbf{99.3} & \textbf{99.4} \\ 

Desk~\cite{roberts2011structure} 
& \underline{89.2} & \underline{91.0} & \underline{92.6} 
& 51.8 & 53.0 & 53.9 
& 88.9 & 90.8 & 92.2 
& 89.0 & 90.8 & 92.2 
& 48.1 & 60.0 & 70.8 
& \textbf{99.1} & \textbf{99.4} & \textbf{99.7} \\ 

Street~\cite{roberts2011structure} 
& 28.3 & 33.3 & 40.5 
& 17.5 & 19.5 & 27.2 
& 28.0 & 43.5 & 67.9 
& \underline{87.4} & \underline{92.4} & \underline{96.2} 
& 20.2 & 28.5 & 39.1 
& \textbf{96.0} & \textbf{97.6} & \textbf{98.8} \\ 

Cup~\cite{roberts2011structure} 
& 29.8 & 37.3 & 43.7 
& 23.6 & 27.2 & 29.9 
& 28.0 & 43.5 & 67.9 
& \textbf{93.6} & \textbf{96.2} & \textbf{98.1} 
& 20.4 & 30.0 & 37.5 
& \underline{66.4} & \underline{78.4} & \underline{87.6} \\ 

Indoor~\cite{jiang2012seeing} 
& 25.7 & 35.7 & 43.9 
& 23.6 & 32.6 & 40.0 
& 55.9 & 73.0 & 86.5 
& \underline{65.2} & \underline{78.8} & \textbf{89.4} 
& 25.6 & 35.6 & 44.0 
& \textbf{68.0} & \textbf{79.9} & \underline{89.3} \\ 

Brandenburg gate~\cite{heinly2014correcting} 
& 29.0 & \underline{46.5} & \underline{67.3} 
& 28.7 & 45.4 & 65.2 
& \textbf{56.1} & \textbf{62.0} & \textbf{67.5} 
& 23.1 & 36.5 & 52.6 
& \underline{41.2} & 45.6 & 50.0 
& 27.0 & 42.8 & 61.6 \\ 

Arc\ de\ triomphe~\cite{heinly2014correcting} 
& 14.6 & 22.8 & 34.2 
& 14.2 & 22.9 & 35.5 
& \underline{57.0} & \underline{61.6} & \underline{65.8} 
& 17.2 & 30.6 & 52.4 
& 20.1 & 22.3 & 24.4 
& \textbf{73.9} & \textbf{80.4} & \textbf{86.3} \\ 

Family~\cite{knapitsch2017tanks} 
& 20.3 & 33.9 & 58.5 
& 81.5 & 88.9 & 94.4 
& \textbf{87.5} & \textbf{92.5} & \textbf{96.2} 
& 19.1 & 32.5 & 56.6 
& \underline{87.1} & \underline{92.2} & \underline{96.1} 
& 80.4 & 88.2 & 94.1 \\ 

Auditorium~\cite{knapitsch2017tanks} 
& 16.5 & 32.9 & 60.0 
& 18.2 & 35.5 & 60.8 
& \textbf{75.9} & \textbf{85.2} & \textbf{92.5} 
& 16.7 & 33.0 & 60.1 
& 14.2 & 15.9 & 17.4 
& \underline{40.6} & \underline{64.7} & \underline{79.4} \\ 

exhibition\_hall~\cite{2017eth3d} 
& 65.4 & 77.5 & 87.2
& 33.6 & 38.8 & 42.9
& 49.0 & 59.6 & 69.6
& 61.0 & 72.6 & 82.0
& 44.3 & 55.6 & 68.2  
& \textbf{70.7} & \textbf{82.1} & \textbf{91.6} \\ 

office~\cite{2017eth3d} 
& 0.0 & 0.0 & 0.0
& \underline{45.1} & \underline{52.7} & \underline{61.2}
& 19.5 & 23.4 & 29.1
& 0.0 & 0.0 & 0.2
& 15.1 & 20.3 & 25.7 
& \textbf{47.8} & \textbf{54.0} & \textbf{61.9} \\ 

old\_computer~\cite{2017eth3d} 
& 6.8 & 9.6 & 13.7
& 42.7 & 50.2 & 57.0
& \textbf{65.6} & \textbf{76.7} & \textbf{85.9}
& 0.3 & 0.6 & 1.7
& 8.5 & 10.7 & 12.1 
& \underline{64.3} & \underline{75.7} & \underline{85.5} \\ 

Campus~(Ours) 
& 42.1 & 45.9 & 48.8 
& 34.5 & 42.5 & 49.3 
& \underline{47.3} & \underline{67.2} & \underline{83.2} 
& 27.9 & 41.6 & 63.8 
& 41.7 & 59.8 & 79.1 
& \textbf{80.7} & \textbf{87.8} & \textbf{93.2} \\

Library~(Ours) 
& 78.5 & 86.4 & 93.2
& 43.2 & 45.7 & 47.7 
& 75.3 & 84.1 & 92.0 
& \textbf{79.8} & \textbf{87.2} & \textbf{93.6} 
& 74.1 & 83.6 & 91.8 
& \underline{78.8} & \underline{86.7} & \underline{93.3} \\ 

Garden-1~(Ours) 
& 14.4 & \underline{29.5} & \underline{54.0} 
& 0.8 & 1.7 & 5.4 
& 7.7 & 11.5 & 14.4 
& \underline{17.9} & 23.5 & 27.9 
& 0.9 & 1.9 & 5.2 
& \textbf{64.4} & \textbf{74.8} & \textbf{84.8} \\

Park~(Ours) 
& 23.4 & 37.1 & 53.5 
& 15.6 & 24.8 & 39.4
& \underline{30.5} & 40.9 & 52.5 
& 29.4 & \underline{44.1} & \underline{59.4}
& 17.7 & 21.6 & 25.1
& \textbf{76.2} & \textbf{83.9} & \textbf{90.8}  \\
\hline
\end{tabular}
}
\caption{\textbf{Comparison with COLMAP~\cite{schonberger2016structure}, GLOMAP~\cite{pan2024glomap}, Doppelgangers~\cite{cai2023doppelgangers}, Doppelgangers++~\cite{Xiang2025dopp++}, GraphSfM~\cite{chen2020graph}, and Ours.} The best results are shown in bold, and the second-best results are underlined.}
\label{tab:comparison}
\end{table*}

\begin{figure}[h!]
\centering
\includegraphics[width=1.0\linewidth]{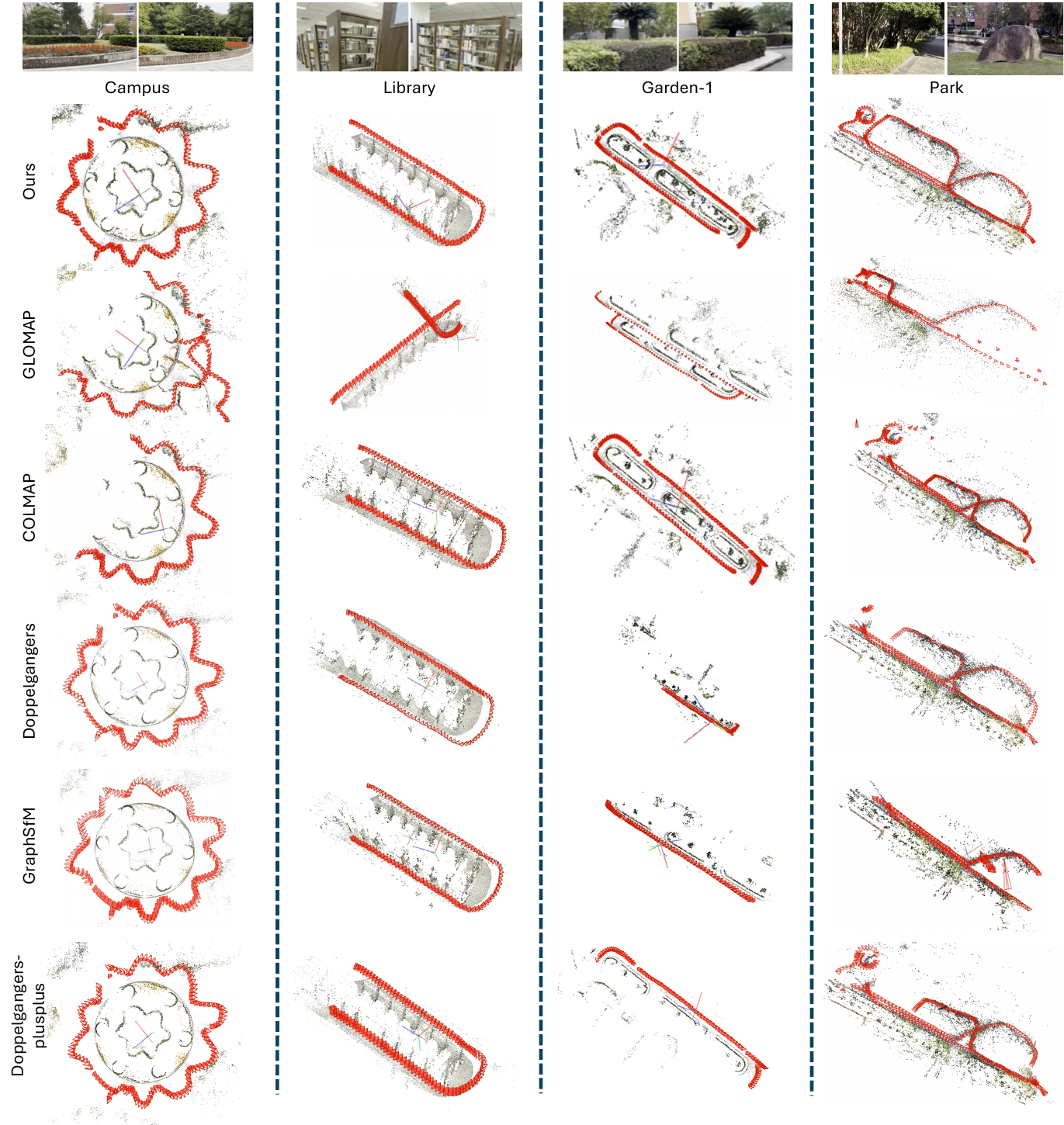}
\caption{\textbf{Qualitative evaluation on sequential datasets.}}
\label{fig:cmp2} 
\end{figure}

\subsection{Evaluation on Sequential Datasets}


We also evaluate our method on sequential image datasets from \cite{knapitsch2017tanks}, covering outdoor and indoor scenes. Specifically, we provide results on the  ``Family'', ``Auditorium'', ``Courthouse'' and ``Meetingroom'' datasets. Again, we compare our method with GLOMAP~\cite{pan2024glomap}, COLMAP~\cite{schonberger2016structure}, Doppelgangers~\cite{cai2023doppelgangers}, Doppelgangers++~\cite{Xiang2025dopp++} and GraphSfM~\cite{chen2020graph}. 
The quantitative evaluation results are presented in Table \ref{tab:comparison}. Our method achieves substantially higher AUC scores than COLMAP\cite{schonberger2016structure} on both Family and Auditorium, although Doppelgangers\cite{cai2023doppelgangers} achieves the best performance on these two sequences. In this dataset, COLMAP~\cite{schonberger2016structure} recovers structures robustly; however, its computational cost is significantly higher than ours, as shown in Table \ref{tab:ablation_comp_time}. Please refer to the supplementary material for additional qualitative comparisons.

We additionally evaluate three sequences from the benchmark of\cite{2017eth3d}: “Exhibition Hall,” “Office,” and “Old Computer.”Our method achieves the best AUC scores on Exhibition Hall and Office and ranks second to Doppelgangers on Old Computer.

We also recorded our own datasets, namely ``Campus'', ``Library'', ``Garden-1'', and ``Park'', as shown in Fig. \ref{fig:cmp2}. For GLOMAP~\cite{pan2024glomap} in Row 2, a significant portion of each scene is missing, primarily because our captured scenes contain large areas with similar appearances. GLOMAP~\cite{pan2024glomap} is relatively sensitive to matching errors caused by these ambiguities. Doppelgangers~\cite{cai2023doppelgangers} uses a network to address this issue, resulting in good performance on the ``Campus'', ``Library'', ``Park'' datasets. However, it misses a large portion of the structures in ``Garden-1''. COLMAP~\cite{schonberger2016structure} recovers the complete structures in most scenes but terminates midway on the ``Campus'' scene, possibly due to false positive edges being selected as the next view. GraphSfM~\cite{chen2020graph} also produces unsatisfactory results for the ``Garden-1'' and ``Park' datasets. The quantitative results provided in Table~\ref{tab:comparison} show that the AUC scores of our reconstruction is generally higher than GLOMAP~\cite{pan2024glomap}. 


\begin{table}[t]
\small
\center
\resizebox{\linewidth}{!}
{
\begin{tabular}{c c c c c c}
\hline
\multirow{2}{*}{Method} & COLMAP & GLOMAP & Doppelgangers & GraphSfM & Ours \\
& PSNR$\uparrow$ & PSNR$\uparrow$ & PSNR$\uparrow$ & PSNR$\uparrow$ & PSNR$\uparrow$ \\
\hline
Books & 17.38 & 17.32 & 16.77 & 17.92 & \textbf{22.63} \\
Cup & 17.61 &	14.59 &	29.78 &	15.85 &	\textbf{31.20} \\
Desk & 17.78 & 22.20 &	24.54 &	24.50 &	\textbf{26.31} \\
Street & 18.51 & 16.89 & 17.74 & 17.69 & \textbf{21.93} \\ 
\hline
\end{tabular}
}
\caption{\textbf{Evaluation using 3DGS.} Our estimated camera poses produce improved novel view synthesis results, demonstrating that our method can produce globally consistent camera poses.}
\label{tab:3dgs}
\end{table}

\begin{table}[t]
\setlength\tabcolsep{2.0pt}
\small
\center
\resizebox{\linewidth}{!}
{
\begin{tabular}{c c c c c c c}
\hline
\multirow{2}{*}{Method} & \multirow{2}{*}{\#Image} & COLMAP & GLOMAP & Doppelgangers & GraphSfM & Ours \\
& & Time & Time & Time & Time & Time \\
\hline
Meetingroom & 371 & 7~min & 3.5~min & 120~min & 13~min & 5.5~min \\
Courthouse & 1106 & 210~min & 30~min & 300+~min & 83~min & 66~min \\
\hline
\end{tabular}
}
\caption{\textbf{Computational time comparison.} The runtime of our method remains substantially lower than COLMAP\cite{schonberger2016structure} and Doppelgangers\cite{cai2023doppelgangers}, while introducing additional computational overhead compared with GLOMAP\cite{pan2024glomap}.}
\label{tab:ablation_comp_time}
\end{table}

\subsection{Evaluation on Unordered Internet Image Datasets}

Moreover, we evaluated our method on unordered Internet image datasets~\cite{heinly2014correcting}. We tested on the ``indoor'' dataset from \cite{jiang2012seeing}, and the ``Brandenburg Gate'' and ``Arc de Triomphe'' from \cite{heinly2014correcting}. Comparisons with GLOMAP~\cite{pan2024glomap}, COLMAP~\cite{schonberger2016structure}, Doppelgangers\cite{cai2023doppelgangers}, Doppelgangers++~\cite{Xiang2025dopp++} and GraphSfM~\cite{chen2020graph} are also included in Table \ref{tab:comparison}. On Indoor, our method achieves the best AUC@3° and AUC@5° and ranks second at AUC@10°, with 89.3 compared with 89.4 for Doppelgangers++. On Brandenburg Gate, Doppelgangers performs best, whereas on Arc de Triomphe, our method achieves the best results across all three thresholds. We include a qualitative comparison in the supplementary materials.

\begin{table}[t]
\centering
\small
\resizebox{\columnwidth}{!}{
\begin{tabular}{c c c c c}
\hline
\multirow{2}{*}{Dataset} &
\multirow{2}{*}{\#Image} &
Full &
W/O Recursive &
W/O $e_{\mathrm{l}}$ \\
& &
AUC@10$^\circ$$\uparrow$ &
AUC@10$^\circ$$\uparrow$ &
AUC@10$^\circ$$\uparrow$ \\
\hline
Garden-1 &
140 &
\textbf{84.8} &
14.6 &
34.7 \\
\hline
\end{tabular}
}

\vspace{2pt}

\resizebox{\columnwidth}{!}{
\begin{tabular}{c c c c}
\hline
W/O $e_{\mathrm{r}}$ &
W/O $e_{\mathrm{t}}$ &
\#RC=20 &
\#RC=40 \\
AUC@10$^\circ$$\uparrow$ &
AUC@10$^\circ$$\uparrow$ &
AUC@10$^\circ$$\uparrow$ &
AUC@10$^\circ$$\uparrow$ \\
\hline
25.9 &
31.2 &
37.3 &
84.8 \\
\hline
\end{tabular}
}
\caption{\textbf{Ablation studies.}
AUC@10$^\circ$ is reported on the Garden-1 dataset.
``W/O Recursive'' denotes performing only a single Louvain partition
without recursively subdividing inconsistent subgraphs.
W/O $e_{\mathrm{l}}$, W/O $e_{\mathrm{r}}$, and W/O $e_{\mathrm{t}}$
denote removing the loop consistency, relative rotation consistency,
and relative translation consistency terms, respectively.
\#RC denotes the number of RANSAC iterations, while the full model
uses 30 iterations.}

\label{tab:ablation}
\end{table}

\subsection{Ablation study}

To validate the effectiveness of the key components in our method, we conduct ablation studies on the recursive subgraph partitioning strategy, the consistency criteria, and the number of RANSAC iterations. The results are reported in Table~\ref{tab:ablation}.

We first evaluate the effect of recursive subgraph partitioning. Removing the recursive partitioning strategy and performing only a single Louvain partition significantly decreases the AUC@10$^\circ$ from 84.8 to 14.6. This result demonstrates that a single graph partition is insufficient to reliably isolate geometrically inconsistent regions. Recursively subdividing subgraphs according to their internal pose consistency is therefore important for obtaining reliable local geometric structures before inter-subgraph edge pruning.

We further evaluate the contribution of the three consistency criteria. Removing the loop consistency term $e_{\mathrm{l}}$
, the relative rotation consistency term $e_{\mathrm{r}}$
, or the relative translation consistency term $e_{\mathrm{t}}$
 decreases AUC@10° from 84.8 to 34.7, 25.9, and 31.2, respectively. The largest degradation occurs when e
 is removed, while all three terms contribute substantially to robust view graph refinement.

Finally, we study the effect of the number of RANSAC iterations in the inter-subgraph edge pruning stage. Reducing the number of iterations from 30 to 20 decreases the AUC@10$^\circ$ from 84.8 to 37.3, indicating that an insufficient number of iterations may fail to identify a reliable consensus among inter-subgraph edges. Increasing the number of iterations to 40 yields the same AUC@10$^\circ$ of 84.8 as the full setting. This suggests that 30 iterations are sufficient on Garden-1 under this evaluation metric, while increasing the number to 40 provides no further improvement in AUC@10°. Therefore, we use 30 RANSAC iterations in all experiments to balance robustness and computational efficiency.


\textbf{Computational time}. Table \ref{tab:ablation_comp_time} provides a comparison of the computational time for all methods. As our approach follows the global SfM paradigm, it is highly time-efficient, particularly for longer sequences. Although our method incurs slightly more time compared to the vanilla GLOMAP~\cite{pan2024glomap}, it still runs over three times faster than COLMAP~\cite{schonberger2016structure} and approximately five times faster than Doppelgangers~\cite{cai2023doppelgangers} on the "Courthouse" dataset. For detailed information on our hardware, please refer to the supplementary materials.

\section{Conclusion}

This paper introduces a robust global SfM method capable of handling large scenes with repetitive or duplicate structures. The main components include a subgraph partitioning strategy to maintain intra-subgraph pose consistency and a RANSAC-based strategy to filter inconsistent edges across different subgraphs.

\noindent\textbf{Limitation and future work.} Our method may still encounter errors when repetitive regions are too large or densely distributed. Additionally, our approach relies on several hyperparameters that may not be optimal. In the future, we plan to incorporate learning-based techniques or advanced matching strategies to overcome these limitations.


\bibliographystyle{CVMbib}
\bibliography{main}



\end{document}